\documentclass[twoside,11pt]{article}

\usepackage{jmlrwcp2e}
\usepackage{url}
\usepackage{amsmath}
\usepackage{enumitem}

\ShortHeadings{Conditional distribution variability measures for causality detection}{Fonollosa}
\firstpageno{1}

\begin{document}

\title{Conditional distribution variability measures for causality detection}

\author{\name Jos\'e A. R. Fonollosa \email jose.fonollosa@upc.edu \\
       \addr \\Universitat Polit\`ecnica de Catalunya. Barcelona Tech.
      c/ Jordi Girona 1-3, Edifici D5\\
      Barcelona 08034, Spain
}

\editor{}

\maketitle

\begin{abstract}
In this paper we derive variability measures for the conditional probability distributions of a pair of random variables, and we study its application in the inference of causal-effect relationships. We also study the combination of the proposed measures with standard statistical measures in the the framework of the ChaLearn cause-effect pair challenge. The developed model obtains an AUC score of 0.82 on the final test database and ranked second in the challenge.
\end{abstract}


\section{Introduction}
\label{sec:introduction}

There is no doubt that causality detection is a task of great practical interest. In a wide sense, attributing causes to effects guides all our efforts to understand our world and to solve any kind of real life problems. There is not, however, a simple and general definition of causality and the topic remains a staple in contemporary philosophy.

The development of analytical methods for detecting a cause-effect relationship in a set of ordered pairs of values also lacks of a universal formal definition of causality. From a pure statistical point of view any bivariate joint distribution can be expressed as the product of any of the two marginal distributions by the conditional distribution of the other variable given the first. And these two equivalent expressions can also be used to explain the generation process in both directions.

In order to be able to attack the causality detection problem we need to introduce one or more assumptions about the generation process or the shape of the joint distribution. Most of those assumptions come from the Occam's razor succinctness principle. We expect to have a simpler model in the correct direction that in the opposite, i.e. the algorithmic complexity or minimum description length of the generation process should be lower in the true causal direction than in the opposite direction. To be more precise, if the random variable $X$ is the cause of the random variable $Y$ we usually expect the conditional distribution $p(Y|X=x)$ to be unimodal or at least to have a similar shape for different given values $x$ of $X$.

Several methods have been proposed in the literature as practical measures of the uncomputable Kolmogorov complexity of the generation model in the hypothetical causal direction. See \cite{Statnikov2012} for a review of the usual assumptions and generation models. In this paper we develop new causality measures based on the assumption that the shape of the conditional distribution $p(Y|X=x)$ tends to be very similar for different values of $x$ if the random variable $X$ is the cause of $Y$. The main difference with respect to previous methods is that we do not impose a strict independence between the conditional distribution (or noise) and the cause. However we still expect the conditional distribution to have a similar shape or similar statistical characteristics for different values $x$ of the cause.

The developed features are combined with standard statistical features following a machine learning approach: the selection of a good set of relevant features and of an adequate learning model.

\section{Features}
In this section we enumerate the features used by our model. All the measures are computed in both directions, i.e., exchanging the role of the two random variables X and Y, except if the measure is symmetric.

\subsection{Preprocessing}
    \textbf{Mean and Variance Normalization.}
Numerical data is normalized to have zero mean and unit variance. All of our features are scale and mean invariant.\\[4 pt]
    \textbf{Discretization of numerical variables.}\label{sec:quantization}
Discrete measures as the discrete entropy and discrete mutual information are also used as features of numerical date after a discretization or quantization process. The quantization uses $2*maxdev*sfactor+1$ equally spaced segments of $\sigma/sfactor$ length and truncates all absolute values above $maxdev*\sigma$. For almost all measures requiring a discretization of the input we selected $sfactor=3$ and $maxdev=3$ in our experiments, i.e, a quantization to 19 different values.\\[4 pt]
    \textbf{Relabeling of categorical variables.}
    \label{sec:sort}
The specific values assigned to categorical data are assumed to have no information by themselves. However, in some cases we considered the calculation of \textit{numerical} measures (as skewness) for categorical variables. For these computations we assigned integer values to the categorical variables as a function of its probability. After the relabeling of variables with M different categories we have: $p(x=0) \geq p(x=1) \ldots \geq p(x=M-1)$. This step let us obtain \textit{numerical} features of categorical variables that do not depend on the labels but on the sorted probabilities.

\subsection{Information-theoretic measures}
In the baseline system we include the standard information-theoretic features as entropy and mutual information. Both the discrete and the continuous version of the entropy estimator are applied to numerical and categorical data after the preprocessing described above.\\[4 pt]
    \textbf{Discrete entropy and joint entropy.}
The entropy of a random variable is a information-theoretic measure that quantifies the uncertainty in a random variable. In the case of a discrete random variable X, the entropy of X is defined as:
	\[H(X) = -\sum_{x} p(x)\operatorname{log}(p(x))\]
In our implementation of the discrete entropy estimator we added the simple \cite{miller1955nbi} bias correction term to finally obtain
	\[\hat{H}_m(X) = -\sum_{x}\frac{n_x}{N}\operatorname{log}(\frac{n_x}{N})+\frac{M-1}{2N}\]
where M is the number of different values of the random variable X in the data set. We also considered the normalized version	$\hat{H}_n(X) = \hat{H}_m(X)/\operatorname{log}(N)$ where $log(N)$ is the maximum entropy a discrete random variable with N different values. The definition and estimation of the entropy can be extended to a pair of variables replacing the counts $n_x$ by the counts $n_{x,y}$ of the number of times the pair $(x,y)$ appears in the sample set.\\[4 pt]
    \textbf{Discrete conditional entropy.}
The conditional entropy quantifies the average amount of information needed to describe the outcome of a random variable Y given that the value of another random variable X is known. In our implementation, the discrete conditional entropy $H(Y|X)$ is computed as the difference between the discrete joint entropy $H(Y,X)$ and the marginal entropy $H(X)$\\[4 pt]
    \textbf{Discrete mutual information.}
The Mutual Information is the information-theoretic measure of the dependence of two random variables. It can be computed from the entropy of each of the variables and its joint entropy as $I(X;Y) = H(X) + H(Y) - H(X,Y)$
In addition to the above unnormalized version, we also included as features two normalized versions. The mutual information normalized by the joint entropy and the mutual information normalized by the minimum of the marginal entropies:
	\[I_j(X;Y) = \frac{I(X;Y)}{H(X,Y)}\hspace{10 mm}I_h(X;Y) = \frac{I(X;Y)}{min(H(X),H(Y)}\]
    \textbf{Adjusted mutual information.}
The Adjusted Mutual Information score is an adjustment of the Mutual Information measure. It corrects the effect of agreement solely due to chance, \cite{Vinh:2009:ITM:1553374.1553511}. This feature is computed with the scikit-learn python package, \cite{scikit-learn}.\\[4 pt]
    \textbf{Gaussian and uniform divergence.}
These features are an estimation of the Kullback-Leibler divergence or \textit{distance} of the distribution of the data with respect to a normalized Gaussian distribution and a uniform distribution respectively. After mean and variance normalization, the estimation of the Gaussian divergence is equivalent to the estimation of the differential entropy except for a constant factor.
	\[D_g(X)=D(X||G)=H(X)-H(G)=H(X) - \frac{1}{2}\operatorname{log}(2\pi e)\]
An estimator of the differential entropy can also be used to compute the divergence respect an uniform distribution if the samples are first normalized in range:
	\[X_u = \frac{X-min(X)}{max(X)-min(X)}\hspace{10 mm}D_u(X) = D(X_u||U)=H(X_u) - H(U)=H(X_u)\]

\subsection{Conditional distribution variability measures.}
In this section we define distribution variability measures that are used as tests of the spread of the conditional distribution $p(Y|X=x)$ for different values of $x$. If this variable is numerical we apply first the quantization process described in \ref{sec:quantization}.\\[4 pt]
    \textbf{Standard deviation of the conditional distributions.}
This is a direct measure of the spread of the conditional distributions after normalization. If $Y$ is a numerical variable, the conditional distribution $p(Y|X=x)$ is normalized for each value of $x$ to have zero mean and then quantized as in section \ref{sec:quantization}. If $Y$ is a categorical variable, the variability of the conditional distribution $p(Y|X=x)$ for different values of $x$ is calculated after sorting these probabilities for each $x$. The standard deviation of the preprocessed conditional distributions is then computed as:
	\[CDS(X,Y) = \sqrt{\frac{1}{M}\sum_{y=0}^{M-1}var_x(p_n(y|x))}\]
where $p_n(y|x)$ refers to the normalized conditional probability and $var_x$ to the sample variance over $x$.\\[4 pt]
    \textbf{Standard deviation of the entropy,  skewness and kurtosis}
These additional features use the standard deviation to quantify the spread of the entropy, variance and skewness of the conditional distributions for different values $x$ of the hypothetical cause
	\[HS(X,Y) = std_x(H(Y|X=x))\hspace{10 mm}SS(X,Y) = std_x(skew(Y|X=x))\]
	\[KS(X,Y) = std_x(kurtosis(Y|X=x))\]
    \textbf{Bayesian error probability}This feature is an estimation of the average probability of error using the (discretized) conditional distributions . For each value of $x$ the probability of error is computed as one minus the probability of guessing $y$ given $x$ if we choose for the prediction $\hat{y}$ the value that maximizes $p(Y|X=x)$. $EP(X,Y) = E[p_e(x)]$ where $p_e(x) = 1-max_y(p(Y|X=x))$

\subsection{Other features}
    \textbf{Number of samples and number of unique samples}\\[4 pt]
    \textbf{Hilbert Schmidt Independence Criterion (HSIC)} This standard independence measure is computed using a python version of the MATLAB script provided by the organizers.\\[4 pt]
    \textbf{Slope-based Information Geometric Causal Inference (IGCI)} The IGCI approach for causality detection, \cite{Janzing20121} proposes measures based on the relative entropy and a \textit{slope-based} measure that we also added to our set of features.\\[4 pt]
    \textbf{Moments and mixed moments} We included the skewness and kurtosis of each of the variables as features, as well as the mixed moments $m_{1,2} = E[xy^2]$ and $m_{1,3}=E[xy^3]$\\[4 pt]
    \textbf{Pearson correlation} The \textit{Pearson r} correlation coefficient computed by the \textit{scipy} python package, \cite{scipy}\\[4 pt]
    \textbf{Polynomial fit} We propose two features based on a polynomial regression of order 2. The first feature is based on the absolute value of the second order coefficient. We have observed that the causal direction usually requires a smaller coefficient. The second feature measures the regression mean squared error or residual.

\section{Classification model selection}

We tested different learning methods for classification and regression. Gradient Boosting, \cite{hastie01statisticallearning}, significantly performed better that the rest of algorithms in our 10-Fold cross-validation experiments on the training set after a manual hyperparameter tuning. We used the scikit-learn implementation (GradientBoostingClassifier) with 500 boosting stages and individual regression estimators with a large depth (9).

The classification task of the ChaLearn cause-effect pair challenge is in fact a three-class problem. For each pair of variables $A$ and $B$, we have a ternary truth value indicating whether $A$ is a cause of $B$ (+1), $B$ is a cause of $A$ (-1), or neither (0). The participants have to provide a single predicted value between $-\infty$ and $+\infty$, large positive values indicating that $A$ is a cause of $B$ with certainty, large negative values indicating that $B$ is a cause of $A$ with certainty, and middle range scores (near zero) indicate that neither $A$ causes $B$ nor $B$ causes $A$.
The official evaluation metric was the average of two Area Under the ROC curve (AUC) scores. The first AUC is computed associating the truth values 0 and -1 to the same class (the class 1 versus the rest), while the second AUC is computed grouping toghether the 1 and 0 classes (the class -1 versus the rest).

Note that the symmetry of the task allow us to \textit{duplicate} the training sample pairs. Exchanging $A$ with $B$ in an example of class $c$ provides a \textit{new} example of the class $-c$.

To deal with this ternary classification problem we tested 3 different schemes:
\begin{enumerate}[leftmargin=*]
	\item A single ternary classification or regression model. The predicted value is computed in this case as $p_1=p(1)-p(-1)$ where $p(1)$ and $p(-1)$ are the estimated probabilities assigned by the classifier to class 1 and class -1 respectively. Alternatively, we can use the output of any regression model. In the case of the selected Gradient Boosting model the classifier version with the \textit{deviance} loss function gave better results than the regressor loss functions in our experiments.
	\item A binary model for estimating the \textit{direction} (class 1 versus class -1) and a binary model for \textit{independence} classification (class 0 versus the rest). The first direction model is trained only with training sample pairs classified as 1 or -1, while the second independence model is trained with all the data after grouping class 1 and -1 in a single class. The predicted value is computed in this case as the product of the probabilities given by each of the models $p_2=p_d(1)p_i(0)$ where $p_d(1)$ is the probability of class 1 given by the direction model and $p_i(0)$ is the independence probability provided by the second model.
	\item A symmetric model based on two binary models. In this scheme we also have two binary models: a model for class 1 versus the rest and another model for class \mbox{-1} versus the rest. In this sense, this configuration follows the same scheme of the evaluation metric. Both binary models are trained with all the training data after the corresponding relabeling of classes. The predicted value is then computed as the difference of the probability given by the first model to class 1 and the probability given by the second model to class -1, $p_3=\frac{1}{2}p_{3,1}(1)-\frac{1}{2}p_{3,2}(-1)$.
 \end{enumerate}
Using the same set of selected features, the three schemes provide similar results as shown in Table \ref{tab:schemes}. The proposed final model uses a equally weighted linear combination of the output of each of the three models to obtain an additional significant gain respect to the best performing scheme.
\begin{table}
	\begin{center}
		\begin{tabular}{|l|l|}
		\hline
		\multicolumn{1}{|c|}{Scheme} & Score \\
		\hline
			1. Single ternary model & 0.81223\\
			2. Direction / Independence models & 0.81487 \\
			3. Symmetric models &  0.81476 \\
		\hline
			System combination  & 0.81960 \\
		\hline
		\end{tabular}
		\caption{Performance of the proposed schemes for the ternary model}
		\label{tab:schemes}
	\end{center}
\end{table}

\section{Results} 
The main training database includes hundreds of pairs of real variables with known causal relationships from diverse domains. The organizers of the challenge also intermixed those pairs with controls (pairs of independent variables and pairs of variables that are dependent but not causally related) and semi-artificial cause-effect pairs (real variables mixed in various ways to produce a given outcome). In addition, they also provided training datasets artificially generated,
\footnote{http://www.causality.inf.ethz.ch/cause-effect.php?page=data}.

The results presented in this section correspond to the score of the test data given by the web submission system of the cause-effect pair challenge hosted by Kaggle. Previous cross-validation experiments on the training set provided similar results. The table \ref{tab:results} summarizes the results for different subsets of the proposed complete set of features. The baseline system includes 21 features:
number of samples(1),
number of unique samples(2),
discrete entropy(2),
normalized discrete entropy(2),
discrete conditional entropy(2),
discrete mutual information and the two normalized versions(3),
adjusted mutual information(1),
Gaussian divergence(2),
uniform divergence(2),
IGCI(2),
HSIC(1), and 
Pearson R(1)
 
\begin{table}[h]
	\begin{center}
		\begin{tabular}{|l|l|}
		\hline
		\multicolumn{1}{|c|}{Features} & Score \\
		\hline
Baseline(21)                                     & 0.742 \\
		\hline
Baseline(21) + Moment31(2)           & 0.750 \\
Baseline(21) + Moment21(2)           & 0.757 \\
Baseline(21) + Error probability(2)  & 0.749 \\
Baseline(21) + Polyfit(2)                  & 0.757 \\
Baseline(21) + Polyfit error(2)         & 0.757 \\
Baseline(21) + Skewness(2)            & 0.754 \\
Baseline(21) + Kurtosis(2)               & 0.744 \\
		\hline
Baseline(21) + the above statistics set (14)   & 0.790 \\
		\hline
Baseline(21) + Standard deviation of conditional distributions(2)                            & 0.779 \\
Baseline(21) + Standard deviation of the skewness of conditional distributions(2) & 0.765 \\
Baseline(21) + Standard deviation of the kurtosis of conditional distributions(2)    & 0.759 \\
Baseline(21) + Standard deviation of the entropy of conditional distributions(2)    & 0.759 \\
		\hline
Baseline(21) + Measures of variability of the conditional distribution(8)                  & 0.789 \\
		\hline
Full set(43 features)                                                                                                  & 0.820 \\
		\hline
		\end{tabular}
		\caption{Results for different subset of the proposed features}
		\label{tab:results}
	\end{center}
\end{table}

A more detailed analysis of the results of the proposed system and of other top ranking systems can be found in \cite{Guyon2014}.

\section{Conclusions} 
We have proposed several measures of the variability of conditional distributions as features to infer causal relationships in a given pair of variables. In particular, the proposed standard deviation of the normalized conditional distributions stands out as one of the best features in our results. The combination of the developed measures with standard information-theoretic and statistical measures provides a robust set of features to address the causality problem in the framework of the ChaLearn cause-effect pair challenge. In a test set with categorical, numerical and mixed pairs from diverse domains, the proposed method achieves an AUC score of 0.82.


\newpage
\acks{This work has been supported in part by Spanish Ministerio de Econom\' ia y Competitividad, contract TEC2012-38939-C03-02 as well as from the European Regional Development Fund (ERDF/FEDER)}

\bibliography{jarfo}

\newpage

\section*{Appendix A. ChaLearn cause-effect pair challenge. FACT SHEET.}
\label{sec:factsheet}


\noindent {\bf Title:} Conditional distribution variability measures for causality detection\\
\noindent {\bf Participant name, address, email and website:} Jos\'e A. R. Fonollosa, Universitat Politècnica de Catalunya, c/Jordi Girona 1-3, Edifici D5, Barcelona 08034, SPAIN. jose.fonollosa@upc.edu, \url{www.talp.upc.edu}\\
\noindent {\bf Task solved:}  cause-effect pairs\\
\noindent {\bf Reference:} Jos\'e A. R. Fonollosa: Conditional distribution variability measures for causality detection. NIPS 2013 Workshop on Causality\\

\noindent {\bf Method:}\\

\begin{itemize}
\item Preprocessing. Normalization of numerical variables. Relabeling of categorical variables
\item Causal discovery. Standard features plus new measures base on variability measures of the conditional distributions $p(Y|X=x)$ for different values of $x$
\item Feature selection. Greedy selection
\item Classification. Gradient Boosting. Combination of three different multiclass schemes
\item Model selection/hyperparameter selection. Manual hyperparameter selection\\
\end{itemize}

\noindent {\bf Results:}

\begin{table}[h]
\begin{center}
\label{tab:table1}
\begin{tabular}{|c|c|c|}
\hline
Dataset/Task & Official score & Post-deadline score\\
\hline
Final test & 0.81052 & 0.81960 \\
\hline

\end{tabular}
\caption{Result table.}
\end{center}
\end{table}

\begin{itemize}
\item quantitative advantages: the developed model is simple and very fast compared to other top ranking models
\item qualitative advantages: it relaxes the noise independence assumption introducing less strict similarity measures for the conditional probability $p(Y|X=x)$.
\end{itemize}
The complete python code for training the model and reproducing the presented results is available at \url{https://github.com/jarfo/cause-effect}. The training time is about 45 minutes on a 4-core server, and computing the predictions for the test test takes about 12 minutes.\\

\end{document}